# Deconstructing Analogy

Mark T. Keane

School of Computer Science & Informatics,

University College Dublin,

Dublin, Ireland

## Abstract

Analogy has been shown to be important in many key cognitive abilities, including learning, problem solving, creativity and language change. For cognitive models of analogy, the fundamental computational question is how its inherent complexity (its NP-hardness) is solved by the human cognitive system. Indeed, different models of analogical processing can be categorized by the simplification strategies they adopt to make this computational problem more tractable. In this paper, I deconstruct several of these models in terms of the simplification-strategies they use; a deconstruction that provides some interesting perspectives on the relative differences between them. Later, I consider whether any of these computational simplifications reflect *the* actual strategies used by people and sketch a new cognitive model that tries to present a closer fit to the psychological evidence.

# 1. Introduction

The late Steve Jobs' was very fond of a computer-bicycle analogy in which he recounted a *Scientific American* article about the relative energy expenditure of different animals in travelling a kilometre. The article reported that people lagged a long way behind the graceful condor and most other species when walking and running but that a person on a bike beat every other animal hands down (or feet down). Jobs' analogy proposed that the *computer was a bicycle for the mind*, conveying lots of rich analogical ideas about mental effort, innovation and radical change.

After three decades of research (Gentner, 1980, 1983; Gick & Holyoak, 1980, 1983; Keane, 1982, 1988), analogy has moved from being viewed as an interesting, but specific, type of creative problem solving to being viewed as a central process in many core cognitive abilities; including relational reasoning, learning, categorisation, induction, and language evolution (Deutscher, 2005; French, 2002; Gentner & Forbus, 2011; Eysenck & Keane, 2010; Keane, 2010; Markman & Gentner, 2001). The corpus of experimental findings on these uses of analogy is supported by a menagerie of models that endeavour to explain the computational basis for analogical processing. All of these models have to solve one core computational problem; namely, that analogical mapping is very computationally expensive, provably NP-Hard, potentially incurring processing 'to the end of time' on reasonably-sized, analogy problems (Falkenhainer, Forbus & Gentner, 1989; Veale & Keane, 1997; see section 2). As such, these models try to simplify various representational and/or processing aspects of the mapping problem – what we will call *simplification strategies* – to make analogising computationally tractable (see section 3). Indeed, an interesting perspective on analogy models can be gained by classifying them in terms of the simplification strategies they adopt (see section 4). Of course, the $64M question is whether any of these simplification strategies are those used by people (section 5). In section 6, I sketch a new cognitive model – called GIBSON -- that processes analogies in a very computationally light fashion, one that perhaps more closely parallels the behaviour we observe in people.

## 2. Computational Level of Analogy & Sources of Complexity

Marr (1982) proposed that cognitive abilities should be characterized at three distinct but interacting theoretical levels; the computational, algorithmic and hardware levels that respectively capture (i) what needs to be computed, (ii) how the computation is achieved, and (iii) the neurological substrate in which this computation is carried out. The computational level does not talk about how the computation is performed, it just sketches those things that *need to be computed* to, for example, solve the problem of analogical mapping (with some minimal representational assumptions). A computational level account of analogy is useful because it allows us to rise above the algorithmic diversity of different specific models (Costello & Keane, 2000; Keane , Ledgeway & Duff, 1994). It is also useful for highlighting the complexity issues that any algorithmic model must solve.

### 2.1 The Computational Level in Analogical Mapping

At the computational level, analogy can be cast as a mapping between two knowledge representations (usually, called the base and target domains) that finds certain one-to-one correspondences between parts of these descriptions (so-called matches). These matches can be between concepts of different representational types; for example, *object~object* matches, *attribute~attribute* matches or *relation~relation* matches. Gentner's (1980, 1983) structure-mapping theory posited that there are two guiding principles in the computation of this isomorphism: parallel connectivity and systematicity. *Parallel connectivity* proposes that matches between relations will enforce matches between the things those relations connect; specifically, if the propositions *love(jim, mary)* and *love(flo, bibi)* are being matched then the *love~love* match will ensure that *jim~flo* and *mary~bibi* are also matched. *Systematicity* in analogical mapping refers to the preference for systematic sets of connected matches over isolated unconnected matches; in part, because they often support inferences by analogy (i.e., parts of the base domain not present in the target). Finally, making a strong representational distinction between attributes [*hot(x), easy(y), red(z)*] and relations [*walk(x, y), run(y, z), touch(mike, mary)*], structure mapping also proposes that analogy is more concerned with *relation~relation* matches than *attribute~attribute* matches.

So, the Jobs' computer-bicycle analogy might create correspondences between people using bicycles and people using computers, between people travelling physical distances by bicycle to far-away physical places and people travelling mental distances with a computer to a new conceptual places and, and so on. We see parallel connectivity in this analogy when we use the matched relations (like *use~use* and *travel~travel*) to enforce object matches between *person~person*, *bike~computer*, *physical-place~mental-place*. We see systematicity at work when people prefer matches that are causally-connected to the *travel* predicates in the analogy. The relative importance of relational matches over attribute matches can arguably be seen in our tendency *not* to expect that the computer should be red or black like the bicycle.

*2.2 Sources of Complexity in Analogical Mapping*

Implicit in the above account are several other what-needs-to-be-computed details that help us elaborate this computational level account, signaling the inherent complexity of the problem; namely, the formation of matching parts and the determination of some overall interpretation for the analogy. On *match formation*, in finding a set of correspondences between the two domains, a space of putative matches (or *match hypotheses*) needs to be computed; for instance, in the computer-bicycle analogy, we probably need to have formed many possible object matches and other relation matches (e.g., *bike~person, person~computer, use~travel, travel~use, physical-place~person* and so on) before one can determine the key matches of the final analogical interpretation. In finding the overall interpretation, one arguably may need to compute two further spaces, a space of partial mappings (*p-maps*) and a space of large-scale, global mappings (*g-maps*, each of which would be alternative interpretations of the analogy). The *p-map space* is a space consisting of small bundles of connected matches – or partial mappings (*p-maps*) – that can be used to build the overall analogical interpretation (generally, sanctioned by parallel connectivity); for instance, in the computer-bicycle case, we might have a number of p-maps for the lower-level relational matches, one p-map might contain [*use~use, person~person, bicycle~computer*] and another p-map would be [*travel~travel, person~person, bicycle~computer*], recognizing that there many be other p-maps that are inconsistent with these ones (like [*use~travel, person~person, bicycle~computer*], [*travel~use, person~person, bicycle~computer*]). Finally, as any two domains may have several overall interpretations for the analogy, we may also need to compute a global-mapping or *g-map space* from which the best interpretation will be chosen: for instance, a g-map that puts the isomorphically-consistent *use~use*-p-map and *travel~travel*-p-map together would be one possible global interpretation of the computer-bicycle analogy. Of course, as we shall see, depending on one's computational approach (symbolic or connectionist) it may well be possible to avoid elaborating all of these data-structure spaces, though most models do have distinct match, p-map and g-maps spaces.

The sources of computational complexity in analogical mapping have long been shown to arise in computing these different spaces (see Falkenhainer et al., 1989, for full analysis). First, finding all possible matches between two representations is generally cast as $O(n^2)$ on the $n$ of the domain elements, as all the elements of the base domain can be matched with all the elements of the target. Second, in an optimal algorithm, to find the best g-map it is necessary to progressively merge all possible consistent p-maps, a step that may well also iterate over previously merged p-maps. Finding g-maps in this way is hugely computationally intensive, with a worse case of $O(n!)$ on the $n$ of the p-maps found. It is these sources of complexity that lead to the provably NP-hard-ness of analogical mapping (Veale & Keane, 1997; Wareham, Evans & van Rooij, 2011), thus presenting several algorithmic challenges for cognitive models of analogy. As we shall see in the next section, various cognitive models adopt simplification strategies to solve these computational problems by either pruning the items in one or all of these spaces or prioritizing those items for consideration. The deeper question, of course, is whether any of the current cognitive models beat the computational problem in the same way that people beat it?

## 3. Simplification Strategies in Analogy Models

At the computational level, we have seen the inherent computational complexity in analogical mapping has to be resolved by the algorithms of the various cognitive models. Indeed, it is possible to classify the models in the literature in terms of the strategy they use to simplify the computational problem faced. In an ideal world, we might have a model -- let's call its MAGIC – that would immediately reduce the match space to just those matches that were in the best g-map (possibly, even skipping over p-map construction). Unfortunately, such magical simplifications have yet to be discovered. In their stead, we have a number of different simplification strategies that prune or prioritize the items in one or more of the three spaces:

- *Reduce Matches*:  Obviously, reducing the match space is one important way to simplify analogical comparison. By pruning the match space, the overall ambiguity in the comparison may be reduced (e.g., one-to-many matches), as might the prolixity of subsequent structures built from these matches (i.e., p-maps, g-maps). Match reduction can be done in a number of ways: (i) *representational restriction*: uses typing in the representation to limit the match space (e.g., only objects can match objects, only predicates of the same –arity can be matched, unsupported attribute-attribute matches may be excluded); (ii) *parallel connectivity restriction*: uses predicate matches to constrain the space of object (or other) matches, so only objects (i.e., arguments to a predicate) that are placed in correspondence by predicate matches are generated, rather than generating all possible object-object matches between both domains [e.g., matching *kill(fido, fiffy)* and *kill(joe, mary)* will match up *fido~joe* and *fiffy~mary*; so, matches like *fido~mary* and *fiffy~joe* will never be produced if not supported by another predicate match]; (iii) *similarity restriction*:  finally, the match space can be reduced by similarity, where the most extreme, commonly-used strategy is to only allow matches between identical predicates, so that gradations of similarity between the predicates are not admitted (like *kill* will only match *kill* and never match *wound*, *murder*, *harm* or *rub-out*).

- *Prioritize Matches*: aside from reducing the match space, one may also prioritize some matches over others for consideration in subsequent processing. That is, a metric can be applied to the set of matches (e.g., its –arity, frequency, higher-order-ness) and then matches can be ranked in terms of this metric, so that some matches are chosen over others to be the focus of processing. By 'prioritization' we mean that there is an explicit computation of a gradient using the metric across the matches in the space; so, for example, a random strategy that just selects a match without scoring matches, would not meet this definition of prioritization.

- *Reduce P-maps:* Matches are combined to form p-maps that, obviously, rapidly increase in numbers, if there is a lot of ambiguity in the match set. P-maps are bundles of matches that are immediately delivered by the parallel connectivity constraint[1]; found by starting at a root match and progressing down through a hierarchy of matches to terminal object-matches. Obviously, if the p-map space can be reduced, in some way, then subsequent processing of g-maps will be eased. However, p-map space reduction is not an option typically used in many models (perhaps because it is hard to assess what to leave in or out at this stage).

---

[1] Note, sometimes a single predicate match along with its supported object-argument matches will be a single p-map though in most cases p-maps will be somewhat larger (cf. Falkenhainer et al., 1989).

- *Prioritize P-maps.* More commonly, some prioritization of p-maps is used whereby a metric is applied to assess their "goodness" (e.g., the number of matches they contain or a systematicity score) and they are rank ordered using this metric to be selected for subsequent processing into g-maps. Again, prioritization assumes computing a metric over the space of p-maps.

- *Prioritize G-maps*: Just as p-maps can be prioritized, so too g-maps can be prioritized using some metric (e.g., size, systematicity scores)[2]. While optimal algorithms, like the original SME algorithm, considered all possible mergings of p-maps to generate a set of alternative g-maps to find the best, more heuristic models often rank-order the p-map set (on some metric) and then merge a subset of these p-maps to find the best g-map.

As we shall see different cognitive models deploy different simpification strategies to deal with the inherent complexity in analogical processing.

*Table 1: Simplification Strategies Used in Seven Cognitive Models of Analogy*

| Model | Reduce Match | Prioritize Match | Reduce P-map | Prioritize P-map | Prioritize G-map |
|---|---|---|---|---|---|
| SME | ✔ | ✔ | - | ✔ | ✔ |
| IAM | ✔ | ✔ | ✔ | ✔ | - |
| SAPPER | ✔ | ✔ | ✔ | ✔ | - |
| CAB | ✔ | ✔ | - | ✔ | - |
| FARGs | - | ✔ | - | ✔ | ✔ |
| ARP | - | ✔ | - | ✔ | - |
| Gibson | ✔ | ✔ | - | - | - |

---

[2] Note, in theory, the line between p-maps and g-maps can be somewhat grey; most g-maps are mergings of several p-maps but, in principle, a single p-map could be a g-map if it was the only and largest bundle of matches in a comparison.

## 4. Classifying Models by Simplification Strategy

Initially, six models are reviewed by classifying them in terms of the simplification strategies they adopt to perform analogical mapping (see Table 1). The models divide strongly along symbolic-connectionist lines that often treat representations quite differently. It is interesting to see how different simplification strategies are realized in different model architectures and the classification sheds some light on the options that exist to realize plausible and parsimonious cognitive models (a point I return to in section 5 after this review).

### 4.1 Structure-Mapping Engine (SME)

SME, the original rendition of the Structure-Mapping Theory (Gentner, 1983), was really the first, well-specified, cognitive model of analogy. Combining the different versions of the SME (Falkenhainer et al, 1989; Forbus, Ferguson & Gentner, 1994; Forbus & Oblinger, 1990) into its most cognitively-plausible greedy-merge version, it is clear that SME uses a combination of representational match-reduction, p-map prioritization and some g-map prioritization. Optimal SME has a worst case $O(n!)$ complexity on the number of p-maps, though its heuristic greedy-merge version does somewhat better, $O(nlog(n))$.

*Match reduction in SME*: Taking two domains that are explicitly typed as relations, attributes, and objects SME reduces the match space by only allowing matches to be formed between elements of the same type. SME will also only match objects that that have been sanctioned as the arguments of matched predicates. As such, the explicit structure of the representation plays a key role in the program's success at finding analogical correspondences, leading to criticisms that the representations are "hand-tailored" (Chalmers, French & Hofstadter, 1992; Forbus et al., 1998; Morrison & Dietrich, 1995). SME also enforces a strong similarity constraint on matching, only allowing identical predicate matching[3]. This further limits the match space in that the program does not have to consider all the possible gradations of similarity that may exist between non-identical predicate matches.

---

[3] SME can also run with a free-for-all rule-set that allows any predicate of the same -arity to match, though obviously this increases complexity. SMErs have also done some work on re-representation to overcome some of the strictness of the identicality constraint (Kurtz, 2005) but, arguably, these methods involve quite conservative representational changes.

*Match prioritization in SME*: Technically, SME could also be said to use match prioritization, in the sense that it scores each match (e.g., object matches get lesser scores than relational matches). However, it does not use this metric to directly select one match over another but rather uses it to score the systematicity of larger bundles of these matches.

*P-map Prioritization in SME:* From this reduced match set, SME builds p-maps out of hierarchically-connected sets of matches*;* so if the proposition *cause(love(M, T), kill(M, T))* matches *cause(love(A,B), kill(A, B))* then these five matches will be grouped as a p-map (n.b., the smallest p-map is a single-predicate match and its object-argument matches). Each of these p-maps is scored for its structural goodness which reflects the number of matches it contains and the connectivity of those matches (Structural Evaluation Score or SES). This metric then allows SME to prioritize the best/largest p-maps for merging purposes, to find the best, global mapping for the analogy.

*G-map prioritization in SME:* As p-maps are merged they become candidate g-maps that are, in turn, may be prioritized by their structural scores. So, in SME's greedy-merge method it takes the highest-scoring p-map and then successively attempts to merge other p-maps with it (if they are consistent with it), rank ordering these p-maps by their scores. In this way, it heuristically finds the best g-map; but it will also find alternative g-maps using the SES scores. So, in one sense, SME prioritizes g-maps during construction of alternative interpretations for the analogy.

*Comment:* So, overall SME succeeds by a clever mix of major restrictions on the initial match space followed by a gentler prioritization of interesting regions of the p-map and g-map spaces to move towards the best (g-map) interpretation of the analogy. As we shall see, many of these simplification strategies are re-used in the other models that followed SME.

### *4.2 Incremental Analogy Machine (IAM)*

IAM (Keane & Brayshaw, 1988; Keane et al., 1994; Keane, 1997) developed an incremental model of analogising to find a plausible explanation for the ease with which people map analogies. Essentially, Keane and colleagues suggested people worked heuristically from the largest connected structure in the base domain, found matches for it and then folded in other matches to it incrementally based on the order in which match-elements appeared in the domains. Using rather simple mapping puzzles, Keane et al. showed that their model paralleled the time-course of analogising in people, suggesting that order mattered for these types of mapping problem (see Keane, 1997).

IAM uses similar representational, match-reduction techniques to SME combined with a form of p-map prioritization to generate is single, best g-map (though it may backtrack to alternative g-maps if the first choice proves to be poor). As such, IAM does not really explicitly juggle alternative g-maps and, therefore, cannot be said to prioritize them in the same way as SME does.

*Match reduction in IAM*: Uses most of the same match-reduction techniques as SME, only matching elements of the same type, using predicate identicality, exploiting the argument-structure of predicates to form matches. In this way, it shows the same sensitivity to the structure of domains and is open to the same representational-criticisms as SME.

*Match prioritization in IAM*: Occurs in two ways, first in promoting the matching of elements in the largest, connected predicate-set in the base (effectively, leading to p-map reduction) and, second, by using the order in which matches occur in the domain representations as a metric for preferring some matches over others; so as the elements in the base are matched with those in the target they are stored in order and those at the front of this set are chosen first.

*P-map reduction and prioritization in IAM:* To the extent that IAM entertains p-maps it performs a simple reduction in the p-map space, by picking the largest p-map. In actual fact, it could also be said that IAM prioritizes p-maps as the matches it prefers using order-of-occurrence are often minimal p-maps (i.e., a relational match and its matched arguments).

*Comment*: IAM's simplification strategies have been shown to produce correct interpretations for many of the classic analogies; that is, most can be solved by heuristically picking the largest structure in the base and building a g-map from it. The impact of order-of-occurrence has been psychologically verified for some analogy problems, but whether it holds more widely is an open question (see Keane, 1997). As such, IAM is quite a light and simple analogical model.

*4.3 SAPPER*

SAPPER (Veale, 1996; Veale & Keane, 1997) is a connectionist model that addresses the computational complexity of analogy by adopting a very different algorithmic approach. It works in a localist, semantic memory of connected nodes with attribute and relational links, in which putative matches (match-bridges) are created between concepts when certain structural regularities hold in the domain representations. SAPPER uses two rules to build its set of match-bridges: (i) *triangulation rule*, which builds a match-bridge between two concepts if they share an identical predicate (e.g., the attribute *red*), completing the triangle of links and (ii) *squaring rule*, which builds a match-bridge forming the fourth side of a square of links (based, in part, on match-bridges previously constructed by the traingulation rule), when three link-sides are found connecting four concepts in the two domains. When an analogy/metaphor is asserted – such as, "Surgeons are butchers" -- parallel activation is passed out along the links in both domains, passing over the match-bridges between the surgeon and butcher domains. When activation cycles settle the highest-activated (scored) matches are read off as the interpretation of the analogy (i.e., the best g-map). Although SAPPER's connectionist architecture differs significantly from the symbolic approach of SME and IAM, it can still be understood as employing a mixture of representational match-reduction, as well as match and p-map prioritization techniques.

Match reduction in SAPPER: Though SAPPER's knowledge representation is a localist semantic memory (its nodes still represent concepts and arcs various predicates) it is clear that SAPPER's bridges are equivalent to the matches of other models. As such, SAPPER reduces the match space using representational typing and predicate identicality in its triangulation rule.  So, for example, certain match-bridges will never be created unless the objects involved share identical attributes.   In SAPPER runs, this appears to have a major impact on match reduction, where comparisons with about 200+ matches/p-maps in SME are reduced to problems with only 18 matches/p-maps in SAPPER. So, SAPPER  avoids a lot of the computational complexity through a combination of its parallel approach and its pruning of the match space. Unfortunately, one side effect of this solution, it that SAPPER could fall foul of the "hand-tailored" criticism; a close inspection of the domain descriptions shows that, perhaps unwittingly, some seem to contain pivotal attribute correspondences on which successful analogising may hinge.

Match prioritization in SAPPER: The match bridges built in SAPPER have 'richness' scores associated with them to indicate their relative importance.  This score may reflect structural aspects (the connectivity of the concepts involved) and other variables. In this way SAPPER has a metric to prioritize one match over others, though this is really only cashed out when the program settles and the final activation levels of matches are read off as the best g-map.

   *P-map reduction and prioritization in SAPPER*: As a knock on effect of math-reduction, it can be said that SAPPER also reduces the p-map space, though it also employes a form of prioritization of p-maps. SAPPER uses very flat domain representations and as such many (though not all) of its p-maps are essentially cases where a single predicate match is the p-map (i.e., the p-map is just a predicate match and its object-match arguments). As these p-map structures can be said to have a 'richness' score, it is reasonable to say that SAPPER uses p-map prioritization. Though, unlike symbolic models, p-maps are not merged but are just the focus of iterative, activation propagation in the network; after which the highest scoring nodes are read off as the best, global mapping.

   *Comment*: SAPPER advances a model that appears to be very computationally efficient in forming analogies. Extensive tests on a large corpus of domains shows that it can find interesting and fruitful analogical correspondences, even when domain boundaries are not explicitly shown. Perhaps its greatest issue is which aspect of the model is responsible for its success; is it successful because of its representations or its match-pruning techniques or its spreading activiation methods.

## 4.4 Connectionist Analogy Builder (CAB)

CAB (Larkey & Love, 2003) report another connectionist implementation that uses constraint satisfaction in a localist, semantic network (Rumelhart, Hinton & Williams, 1986). It uses predicate identicality to establish match links between the nodes of the two domains and then uses a coding of the directionality of link-paths between nodes to implement parallel connectivity and systematicity. On the face of it, it appears to be a fairly-straight parallel rendition of SME's constraints, with some additional parameters that allow for the modelling of working memory limitations. CAB uses the same representational match-reduction techniques as SME and then uses its iterative constraint-satisfaction methods to prioritize matches to generate a single, best g-map. As CAB uses structural constraints from p-maps, in one sense, it could be said to prioritize them to find the best analogy interpretation; but, this is a moot point given the parallel nature of the model.

*Match reduction in CAB*: Uses most of the same match-reduction techniques as SME, only matching elements of the same type, using predicate identicality, exploiting the argument-structure of predicates to form matches. In this way, it shows the same sensitivity to the structure of domains and is open to the same representational-criticisms as SME.

*Match and p-map prioritization in CAB*: Occurs by promoting the matches in the larger p-maps over others, to be selected as part of the g-map.

*Comment*. CAB is another interesting architectural twist on the SME-story. Like SAPPER it appears to compute analogies quite efficiently, though its complexity is quite sensitive to the structure of the domain.

### 4.5 Fluid Analogies (FARG)

The analogy models developed by the Fluid Analogies Research Group (FARG; Hofstadter & FARG, 1995) represent an important and novel departure from the previous models, primarily because they have much more representational flexibility than the others. In these models, the domain representations are built at the same time as the analogy is being mapped (hence, the high-level perception debate between them and SMErs; Chalmers et al, 1992; Morrison & Dietrich, 1995). For example, in the Copycat model, which developed analogies for letter sequences – how should one complete **abb?** by analogy given **lmmnnn** – one could propose **abbc** or **abbccc** as completions by analogy based on whether one categorises **nnn** as "next-letter-in-alphabetic-sequence" or "next-letter-plus-one-more-than-previous-group". FARG models entertain alternative descriptions of the elements in the domains while forming the analogy, rather than working off given, fixed-predicate descriptions. So, in one sense, these models expand the match space to represent alternative categorisations/representations of the objects involved and then explore these much richer representations, using parallel methods, to really "create" the analogical correspondences rather than finding them "ready-made" via an identicality criterion.

The complexity entailed by the FARG family of models is unclear as they have been largely applied in quite restricted domains (some would say "toy domains"). What is clear, is that they require significant categorisation and solution-goodness knowledge to drive the analogy process[4]. Clearly, there is no real match reduction (as it is often expanded), though there is prioritization of matches/p-maps/g-maps as the model moves progressively to the global interpretation of the analogy. Exactly what type of structure *is* prioritized may be a matter of interpretation, given the parallel nature of the model.

*Match prioritization in FARG*: FARG models clearly prioritize some matches over others, in that a given match can have a different weight (goodness score) in the overall comparison being made (e.g., the "next-letter-in-alphabetic-sequence" mapping may be weighted lower than "next-letter-plus-one-more-than-previous" mapping because the latter covers more attributes of the elements).

*P-map/G-map prioritization in FARG*: While it is a matter of interpretation exactly what is prioritized in these models as they iteratively add categorisations and mapping links between concept nodes, it is probably most reasonable to cast these as progressive g-maps, that are successively prioritized and refined by interactions between the added category information, the weights on mapping links, the competition between alternative mappings and the assessment of the overall goodness of the analogy.

*Comment*: FARG models represent a very significant departure for models of analogy, as they propose a much more dynamic perspective on the building of representations; the analogy emerges from a dialogue between the alternative possible categorisations of the elements of both domains and the global requirements of the comparison. However, unlike many of the previous models, it is less clear what general principles can be applied to other cases of analogy; for instance, concretely how they might be applied to the benchmark analogies regularly discussed in the psychological literature?

### 4.6 Analogical Relational Priming (ARP)

The Analogical Relational Priming model (ARP; Leech, Mareschal & Cooper, 2008) is another connectionist model that uses distributed representations of predicate relations in a recurrent network, where the analogy gradually emerges as a by-product of pattern completion. ARP concentrates on explaining the developmental phenomena found using simple, proportional analogies, such as *apple:cut-apple::*

---

[4] To such a degree that it is hard to imagine how they would operate in the sorts of domains used in the previous analogy models, without modeling significant parts of human knowledge.

*bread: cut-bread*. The basic idea is that relations are encoded in memory as transformations of one state into another (e.g., an *apple* state to a *cut-apple* state) and when presented with a partial analogus domain, the tranformational state is projected onto the new object to generate the analogy (see Keane, Smyth & O'Sullivan, 2001, for a related approach to similarity). ARP is also demonstrated with one larger analogy – the WWII-Gulf War analogy – where it incrementally finds analogical matches, using a random-selection strategy.

So, like the FARG models, in one sense ARP expands the match space in using all the distributed representations associated with the relations and objects in both domains. When analogies get larger than proportional ones, it uses a random strategy to select matches incrementally.

*Match prioritization in ARP*: A random selection strategy is an interesting simplification strategy though it is doubtful whether it would work adequately in larger analogies (n.b., strictly speaking it is not prioritization).  However, Leech et al. do acknowledge that some sort of prioritization of matches might emerge from the top-down structure of domains,  though this idea remains to be cashed out.

*Comment*: In the *Behaviour & Brain Sciences* commentary on Leech et al's paper, many criticisms are elaborated. For example, Holyoak & Hummel (2008) say it throws away all the essential aspects of analogising in handling larger analogies, others question how the model can find analogies between distant domains, where a shared basis for pattern completion would not exist (Dietrich, 2008; French, 2008). One would also worry about the adequacy of the representation of the base relations, in terms, of whether they reflect the true range and diversity of event instances for a given relation.

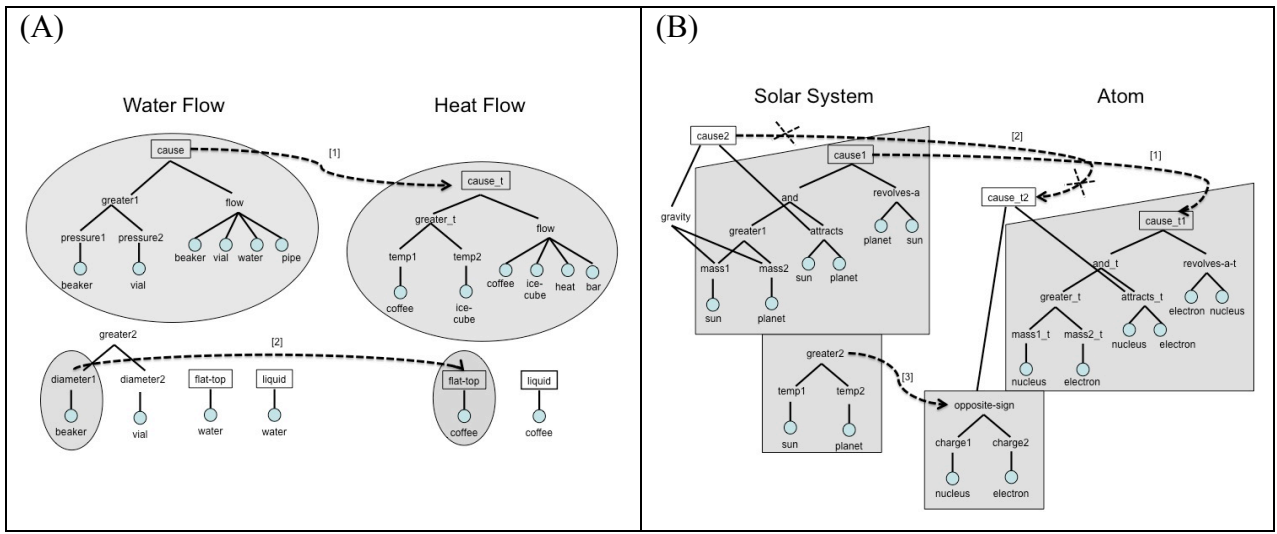

**Figure 1:** *Two analogy problems -- (A) Flow Analogy and (B) Atom Analogy – showing (i) the objects in the domains (blue circles), (ii) the predicates taking part in root matches (predicate names with boxed outlines), (iii) the BestMap initially found by Gibson (shaded areas in both domains), and (iv) the order in which BestMatches were selected for the mapping (numbered, dashed arrows).*

## 5. Direct Psychological Evidence for Simplification Strategies

While cognitive models often assume many hypothetical representations and processes to explain the psychological evidence, it is noteworthy that there is little direct evidence for most of the simplification strategies used in analogy models. For example, in their review of the benchmark psychological effects in the analogy literature, Markman & Gentner (2000) list three main phenomena: (i) that cognitive representations are structured and the comparison process operates to align two structures, (ii) analogical relatedness depends on semantic commonalities between relations in the two domains being compared (iii) the comparison process is driven by a search for correspondences that preserve connections between representational elements (see p. 501). Notably, all the models reviewed here meet these benchmarks, even though many use very different simplification strategies. So, which model best parallels what people actually do?

There are actually very few results in the literature that directly support a given simplification strategy. For instance, there is little evidence to show that certain matches are *never* entertained during analogical comparisons, that the time course of analogizing is hindered by the presence of competing g-maps or that merging p-maps is eased by reduced ambiguity in the match space. To show that such tests are possible consider a few instances. Keane (1997) provided evidence on the time course of mapping using simple mapping problems, showing that order effects occurred in finding mappings; by moving the relative positions of elements in the domain descriptions. Larkey & Love (2003) report that working memory limitations affect particular cross-mapped analogies mapping adopted between two domains when interpretations based on attributes or relational structure are provided. Finally, O'Toole & Keane (2012) found some evidence that comparing objects using different similarity instructions (e.g., "Are these two objects similar?" versus "Are these two objects similar by virtue of their colour?") affected subsequent visual search where an object had to be found in a camouflaged background; indicating that the matches stored from the similarity judgment were different in both cases.

When one considers the unresolved issues around representation, the complexity that still persists in handling large domains, and the lack of direct support for different simplification strategies (and thus inherent lack of parsimony in explanations), one wonders whether it might be worthwhile to develop a minimalist model that might do such things more directly. In the next section, we outline a new model that meets some of these constraints and report some initial results from it.

*Table 1: Gibson's Performance on Two Analogies*

| Measure | Flow | Atom |
|---|---|---|
| Total Matches | 44 | 35 |
| No. of BestMaps | 2 | 3 |
| BestMap produced | 1st | 1st |
| BestMap Matches N | 10 | 12 |
| Cycles to Optimal | 2 | 3 |
| Runtime (secs) | 0.037 | 0.039 |

## 6. GIBSON: A Parsimonious Model of Analogy

Earlier we said that a fictional best model of analogy – we called it MAGIC – would restrict the match set to just those matches that supplied the minimal set of p-maps that merged to deliver the best g-map. This is what the GIBSON model tries to do (Keane, 2012a, 2012b). Like most previous models, GIBSON partially restricts the match space using representational typing (predicates only match predicates, object only match objects) but beyond that it can entertain all manner of matches (including, in theory, alternative classifications of the objects it the domain[5]).

Essentially, GIBSON tries to jump from the matches directly to the subset of matches that make up the best interpretation of the analogy (i.e., the optimal g-map or BestMap); it tries to do this by using the information present in the matches alone (information that can be derived from the original, presented domain representations). In this way, it only constructs p-maps from selected matches when it is merging them with other p-maps to build this global g-map; so, it tries to jump from matches directly to the best interpretation, though when its heuristic guesses fail or a tie between matches is encountered, it can fork its processes to pursue alternative g-maps. So, the GIBSON algorithm essentially prioritizes the match set using a variety of features of the matches generated; essentially, it tries to assess each match for its BestMap Potential, its potential for being one of the matches making up the optimal g-map. At present, in GIBSON, BestMap Potential is based on a formula applied to a match, *M*, between any base element, *b*, and any target element, *t*:

(1)   *BestMapPotential(b, t, M) =*

$\quad$ *|AM(b, t, M)| + Min(L(b), L(t)) + Freq(b) + Freq(t)*

$\quad$ *+ R(b, t, M) + |NN(b, t, M)| * |KN(b, t, M)|*

where it finds the sum of the size of the set of argument matches of the match (*AM*), the minimum of the levels (L) of both elements (i.e., where they are in the hierarchies of the domain), the frequency of the match's elements (*Freq*) in their respective domains, the rooted-ness of the match (*R*) in their respective domains and, finally, the product of the sizes of the set of new matches (NM) added to known matches and the known matches (KM) in the g-map being built. GIBSON iteratively selects matches with highest score until either no matches are left to be checked or all the base elements have been mapped (see Keane, 2012a, for details).

Interestingly, this approach can be shown to generate the correct best mappings for the benchmark analogies in the literature in a very computationally light way (see Figure 1 and Table 1). It also may well be the first analogy model to demonstrably handle very large analogies (e.g., up to 91,000 matches) in reasonable time limits. Table 2 shows GIBSON runs on self-mapped domains (mapping a base domain to itself) from Veale's (1996) professions domains and predicate descriptions of student essays. These runs also illustrate how more natural domains (the student essays) seem to have very different structural properties to ones have have been assembled by researchers (the professions domains).

---

[5] Though this feature is not yet implemented, in any way.

*Table 2: Gibson Median Runtimes and Size Statistics for Several Large Analogies*

| Comparison | Total Matches | Partial Maps | G-map Size | % Corr | Time (secs) |
|---|---|---|---|---|---|
| | | *Literal* | *Similar* | *Rules* | |
| **Professions:** | | | | | |
| Composer-Self | 11,605 | 4,881 | 201 | 63 | 10 |
| General-Self | 49,013 | 29,629 | 407 | 62 | 223 |
| *Essays:* | | | | | |
| Design-Self | 2,493 | 38 | 181 | 100 | 0.57 |
| Compress-Self | 6,326 | 59 | 328 | 100 | 2.49 |
| | | *Free* | *Forall* | *Rules* | |
| **Professions:** | | | | | |
| Composer-Self | 20,885 | 14,161 | 201 | 44 | 26.6 |
| General-Self | 91,145 | 71,824 | 407 | 55 | 628 |
| *Essays:* | | | | | |
| Design-Self | 4,443 | 64 | 181 | 100 | 0.74 |
| Compress-Self | 14,320 | 129 | 328 | 100 | 7.6 |

## 7. Conclusions

In this paper, we have reviewed seven models of analogical mapping in terms of the simplification strategies they adopt. This taxonomic framework provides a handy way of appreciating some of the similarities and differences between the models.  It also points up the extent to which many key aspects of the models lack any direct evidential foundation in the empirical literature (a separate issue that perhaps needs to be addressed).  Lastly, we have introduced a new model of analogy, called GIBSON, that tries work more directly and efficiently and reports on some of its processing parameters.